\documentclass[letterpaper, 10 pt, conference]{ieeeconf}  

\IEEEoverridecommandlockouts                              

\usepackage[pdftex]{graphicx}
\usepackage[inkscapeformat=png]{svg}

\usepackage{amsmath,amsfonts,amsthm,amssymb}  
\usepackage{xfrac}
\usepackage{hyperref}
\usepackage[english]{babel}
\usepackage{mathtools}
\usepackage{algorithmic}
\usepackage[caption=false, font=footnotesize]{subfig}
\usepackage{mwe}
\usepackage{cite}
\usepackage{bm}
\usepackage{wrapfig}
\usepackage[ruled,vlined]{algorithm2e}

\usepackage{etoolbox}
\newtoggle{arxiv}
\toggletrue{arxiv}

\newcommand{\ale}[1]{\textcolor{red}{[Alejandro: #1]}}
\newcommand{\off}[1]{}
\newcommand{\anil}[1]{\textcolor{magenta}{[Anil: #1]}}

\newcommand{\ignore}[1]{}

\def\bx{\bm{x}}
\def\bu{\bm{u}}
\def\bz{\bm{z}}
\def\beps{\bm{\epsilon}}

\theoremstyle{definition}

\theoremstyle{theorem}

\def\niceparagraph#1{\vspace{5pt} \noindent \textbf{#1}}

\def\kiril#1{\textcolor{blue}{(\textbf{Kiril:} #1)}}

\renewcommand{\QED}{\QEDopen}

\title{\LARGE Bringing Network Coding into Multi-Robot Systems: \\ Interplay Study for Autonomous Systems over Wireless Communications 
\vspace{-7pt}
}

\author{Anil Zaher, Kiril Solovey, and Alejandro Cohen
\thanks{The authors are with the Viterbi Faculty of Electrical and Computer Engineering, Technion--Israel Institute of Technology, Haifa, Israel. email:  
{\tt anil.zaher@campus.technion.ac.il, \{kirilsol,alecohen\}@technion.ac.il}.}%
\vspace{-0.8cm}}

\begin{document}

\maketitle
\thispagestyle{empty}
\pagestyle{empty}

\begin{abstract}
Communication is a core enabler for multi-robot systems (MRS), providing the mechanism through which robots exchange state information, coordinate actions, and satisfy safety constraints. While many MRS autonomy algorithms assume reliable and timely message delivery, realistic wireless channels introduce delay, erasures, and ordering stalls that can degrade performance and compromise safety-critical decisions of the robot task. 
In this paper, we investigate how transport-layer reliability mechanisms that mitigate communication losses and delays shape the autonomy–communication loop. We show that conventional non-coded retransmission-based protocols introduce long delays that are misaligned with the timeliness requirements of MRS applications, and may render the received data irrelevant.
As an alternative, we advocate for adaptive and causal network coding, which proactively injects coded redundancy to achieve the desired delay and throughput that enable relevant data delivery to the robotic task. Specifically, this method adapts to  
channel conditions between robots and causally tunes the communication rates via efficient algorithms.

We present two case studies: cooperative localization under delayed and lossy inter-robot communication, and a safety-critical overtaking maneuver where timely vehicle-to-vehicle message availability determines whether an ego vehicle can abort to avoid a crash. Our results demonstrate that coding-based communication significantly reduces in-order delivery stalls, keeps cooperative-localization accuracy close to the ideal baseline when paired with delay-aware re-estimation, and satisfies the overtaking abort deadline in about $80\%$ of the simulated runs, compared with about $60\%$ for the retransmission-based baseline.
Overall, the study highlights the need to jointly design autonomy algorithms and communication mechanisms, and positions network coding as a principled tool for dependable multi-robot operation over wireless networks. Code: \href{https://github.com/AnilZaher/network-coding-mrs}{github.com/AnilZaher/network-coding-mrs}.
\end{abstract}


\section{Introduction}\label{sec:intro}

\begin{figure}[t]
    \centering
    \includegraphics[trim=0.6cm 0.6cm 0.6cm 0.6cm,clip,
        scale=0.3,
        angle=-90,]{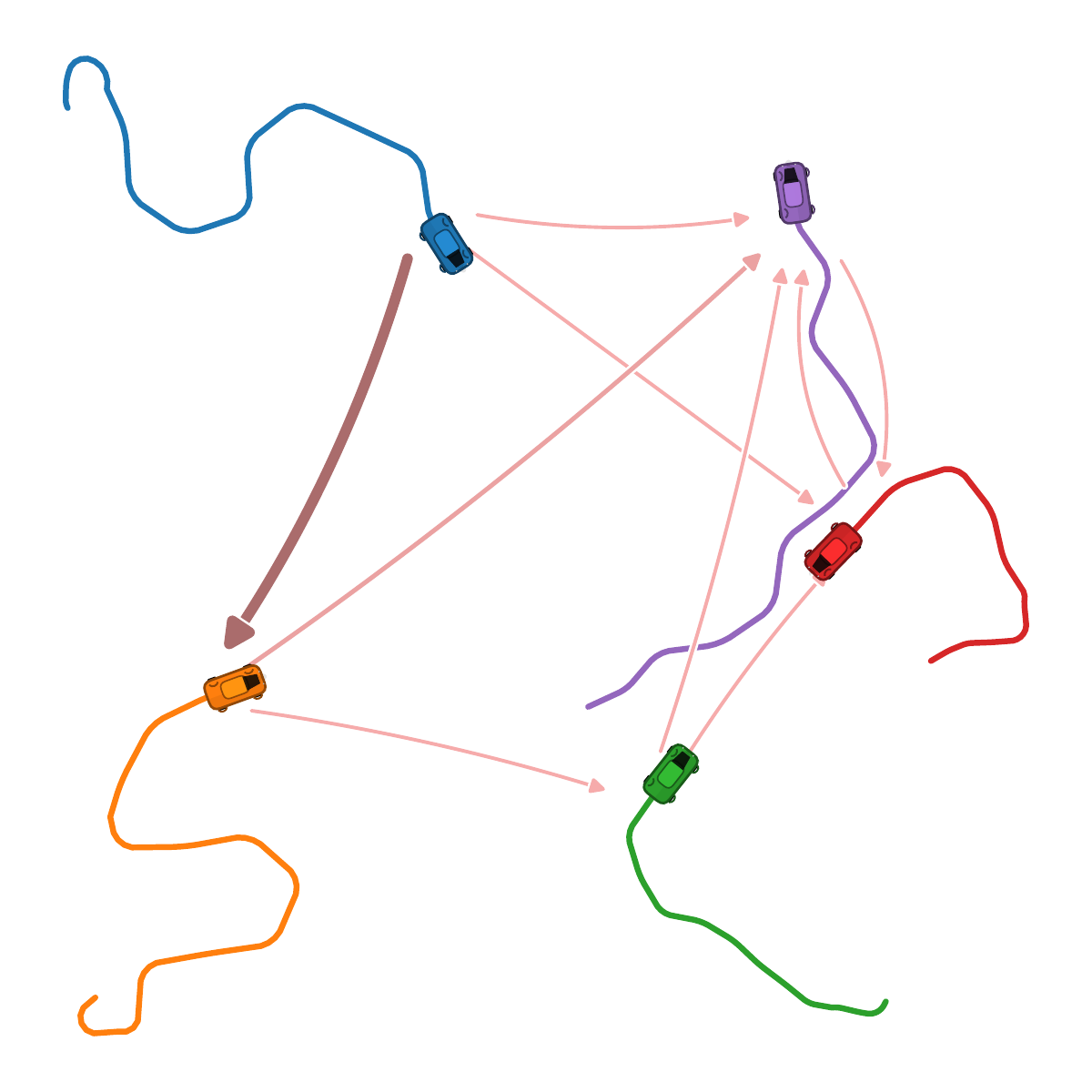}\vspace{0pt}
    \caption{A cooperative localization scenario considered in our first case study (Sec.~\ref{subsec:coopL} and~\ref{subsec:cooploc_results}). 
    Each robot obtains local GPS 
    measurements and inter-robot measurements of nearby robots, which are shared over inter-robot communication channels. The colored curves show the robots' ground-truth trajectories, and the arrows depict the inter-robot observations communicated. Arrow thickness and color intensity encode the communicated measurement delay (darker and thicker arrows indicate larger delays).}
    \label{fig:system_view}
    \vspace{-20pt}
\end{figure}

Distributed systems form the backbone of modern computation, enabling numerous independent computing units to work together as a cohesive whole to achieve shared goals. When these principles are extended into the physical domain, they form the foundation of multi-robot systems (MRS), where networks of mobile agents cooperate to accomplish complex objectives \cite{10433886}.
Recent work demonstrates the potential of such cooperation at scale. For instance, by exchanging information beyond onboard sensing, connected autonomous vehicles can improve traffic efficiency and safety through coordinated maneuvers, cooperative perception, and intersection management \cite{Stern2018TRC,wu2021flow, NHTSA2014V2V,Dresner2008JAIR}. 

Enabling such cooperation places stringent demands on communication. In many coordination and safety-critical applications, state updates must be delivered within tight latency bounds, ranging from tens to hundreds of milliseconds for cooperative awareness, and down to single-digit milliseconds for coordinated maneuvers, while meeting very high reliability targets \cite{8410403,3gppTS22186Rel18}. These requirements are commonly captured under the notion of ultra-reliable low-latency communication (URLLC) \cite{Popovski2019URLLC}, which characterizes the reliability and timeliness needed to support dependable autonomous operation.



These latency and reliability requirements arise directly in cooperative multi-robot applications,  
wherein each robot continuously tracks state information and makes real-time decisions based on information received over wireless links from other robots, while usually assuming timely and reliable delivery. In practice, however, communication networks are inherently non-ideal: wireless interference, shared-medium contention, buffering, and time-varying connectivity introduce delay, packet loss, and timing variability, causing information to arrive late, out of order, or not at all. Such effects can violate the assumptions underlying estimation and decision-making algorithms, potentially degrading synchronization, estimation accuracy, and safety.

\begin{figure*}
\vspace{5pt}
    \centering
\includegraphics[trim={4cm 2.8cm 3cm 2.8cm},clip,width=\iftoggle{arxiv}{0.9}{0.6}\textwidth]{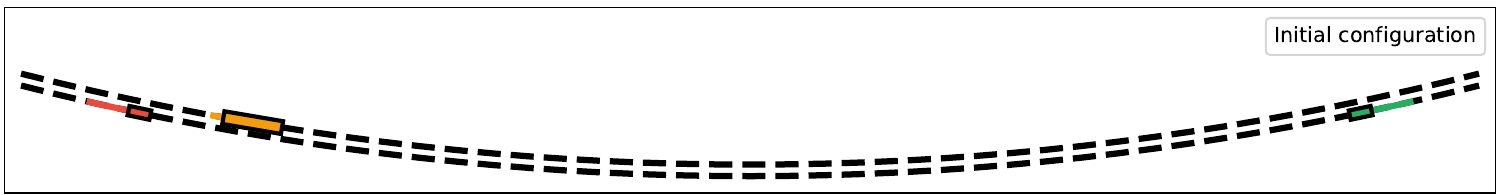} \hfill
\includegraphics[trim={4cm 2.8cm 3cm 2.8cm},clip,width=\iftoggle{arxiv}{0.9}{0.6}\textwidth]{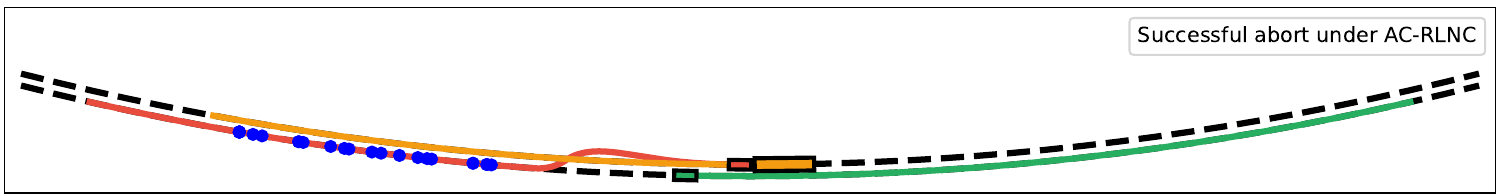}  \hfill
\includegraphics[trim={4cm 2.8cm 3cm 2.8cm},clip,width=\iftoggle{arxiv}{0.9}{0.6}\textwidth]{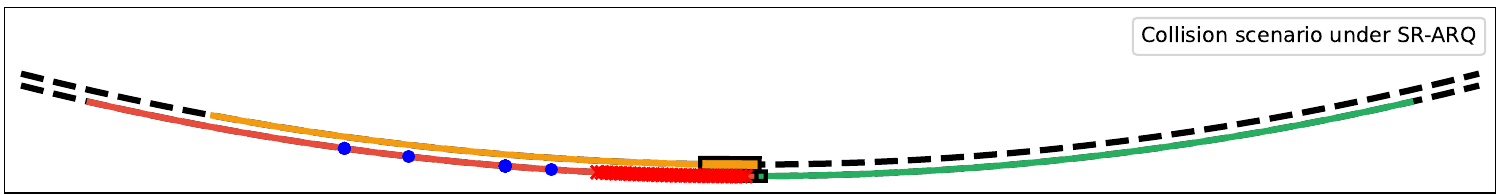} 
    \caption{Visualization of the overtaking scenario we consider in our second case study (see Sections~\ref{subsec:overtake_model} and~\ref{subsec:overtake_results}) with an initial configuration, and two separate runs using AC-RLNC and SR-ARQ transport mechanisms, respectively. Ego vehicle $A$ (red) follows the outer lane behind truck $T$ (yellow), while oncoming vehicle $B$ (green) approaches from the opposite direction on the same lane. Timely V2V packet reception is required for $A$ to detect the oncoming hazard and abort. Blue dots indicate time instants at which V2V packets from vehicle $B$ are successfully received by the ego vehicle $A$. 
    }\label{fig:overtake_overview} \vspace{-17pt}
\end{figure*}

This gap is visible in two representative application areas: cooperative localization and safety-critical vehicle-to-vehicle (V2V)-enabled maneuvers.
Cooperative localization is a central capability in multi-robot systems, where inter-robot measurements improve pose estimation accuracy across the team \cite{Roumeliotis2002Distributed}. Many formulations assume ideal or near-synchronous communication, even though wireless channels routinely introduce delay and packet loss that can destabilize Kalman filter (KF) based estimators \cite{Sinopoli2004IntermittentKF}. Methods that address delayed or out-of-sequence measurements, such as smoothing approaches 
and out-of-sequence measurement (OOSM) 
filtering \cite{BarShalom2002OOSM,Zhang2022RHFLS}, provide mechanisms for incorporating late information, yet they do not examine how communication-layer behavior produces the delay and loss patterns that shape estimator performance. As a result, the interaction between communication-layer behavior and cooperative localization remains insufficiently characterized.


Similarly, V2V communication enables safety applications such as forward-collision warning and cooperative maneuvers \cite{NHTSA2014V2V}. Formal safety frameworks, e.g.,
\cite{Shalev2018RSS}, define timing envelopes for safe decisions, and overtaking assistance systems further illustrate how timely inter-vehicle updates reduce collision risk \cite{article}. However, these studies typically either assume that the communication layer satisfies required timing and reliability bounds or evaluate it only through aggregate metrics, and do not evaluate how protocol behavior influences meeting safety deadlines.


Despite the central role of communication in enabling cooperation, much of the existing work on multi-robot systems treats communication as a secondary concern rather than a co-designed component of the autonomy pipeline. In particular, reliability is often delegated to retransmission-based transport protocols that recover specific lost packets after feedback. While such mechanisms are effective in traditional data networks, they introduce round-trip-time (RTT)  recovery delays and head-of-line blocking\footnote{RTT is the time between transmitting a packet and receiving its acknowledgment; after a loss, retransmission-based protocols incur at least one RTT before recovery. Head-of-line blocking occurs when ordered delivery prevents correctly received packets from being released because an earlier packet is missing.}, which can significantly increase in-order delivery delay under loss. 
This gap has been noted in prior work and motivates tighter co-design between autonomy algorithms and communication mechanisms~\cite{Gielis2022Review}.

\niceparagraph{Contribution.}
Ignoring the communication layer can turn an otherwise correct MRS autonomy pipeline into a brittle system: late, missing, or irregularly-timed information can corrupt estimation and can undermine safety-critical decisions. 
To make these consequences concrete and motivate communication-aware co-design, we present a focused case study that examines the autonomy–communication loop from two complementary perspectives. 
 %
%
    First, we use cooperative localization under practical communication conditions (Fig.~\ref{fig:system_view}) as a case study to examine how transport-layer delivery behavior affects estimation. We compare a basic EKF-based approach with a \emph{delay-aware iterative re-estimation method} (I-ReE), inspired by smoothing and OOSM handling \cite{BarShalom2002OOSM,Zhang2022RHFLS}, under delayed and lossy communication, and then quantify how delivery behavior under non-ideal links, ranging from an unreliable baseline to reliable communication protocols, including adaptive coded protocols, shapes the information available to the estimator and the resulting estimation accuracy.
    Second, we develop a similar analysis for  a safety-critical overtaking scenario (Fig.~\ref{fig:overtake_overview}), where timely V2V message availability determines whether an ego vehicle can safely abort before a computed deadline. 
    
    We examine the complete autonomy and communication loop, rather than evaluating each ingredient in isolation. 
    Overall, we emphasize 
    that communication-layer behavior must be accounted for as a co-designed component of the MRS autonomy pipeline. 
    In addition, we show that the structure of the transport protocol fundamentally shapes the information available to the autonomy pipeline. 
    
    Rather than relying solely on reactive retransmissions, we investigate adaptive causal random linear network coding (AC-RLNC) as a reliability mechanism that proactively injects coded redundancy, which enables decoding based on accumulated degrees of freedom rather than specific packet identities\iftoggle{arxiv}{~\cite{9076631,dias2023sliding,vasudevan2026revisiting}}{~\cite{9076631}}. Our results demonstrate that network coding is not merely a throughput enhancement technique, but a mechanism that can directly affect estimation consistency and safety in multi-robot systems.


\niceparagraph{Organization.}
In Sec.~\ref{sec:background}, we provide necessary background on communication. In Sec.~\ref{sec:sysmodel}, we describe the system models and the two case studies: cooperative localization under communication constraints and a safety-critical overtaking scenario under timing constraints. In Sec.~\ref{sec:experiments}, we present the experimental setup and results for both case studies. We conclude and discuss future directions in Sec.~\ref{sec:conclusion}.
\section{Background on Communication Systems}
\label{sec:background}
This section provides the background for interpreting our case studies and experimental results. It briefly reviews the relevant communication layers, introduces the transport protocols considered herein, and summarizes key communication challenges in connected MRS.


Modern communication networks are often described using a five- or seven-layer architecture, in which each layer provides a distinct set of services to the one above it \cite{1457043}. The following layers are most relevant to our setting.
The \emph{physical layer} governs signal transmission. In MRS, it mainly determines which robot pairs can sustain a usable link given their positions, obstacles and occlusions in the scene, and local propagation conditions (e.g., path-loss, fading, and multipath). The \emph{link layer} organizes
transmissions into frames and arbitrates access to the shared channel; in robot teams it
enables one-hop exchanges among nearby robots over the same wireless channel. The
\emph{network layer} (e.g., Internet Protocol (IP)) provides addressing and routing across multi-hop networks;
in mobile teams it determines how messages are forwarded when robots are not directly
connected and when relays are needed as connectivity changes. The \emph{transport layer}
defines end-to-end delivery behavior (e.g., best-effort vs.\ reliable and unordered vs.\ ordered delivery),
shaping how information is delivered to the application. In robotics, it determines whether autonomy modules receive independent datagrams or delivery with recovery and ordering. 
Finally, the \emph{application layer} exposes
communication services to higher-level algorithms; in robotic systems, it corresponds to task-level
messages such as state/intent broadcasts and shared observations. 


Reliable communication performance is typically characterized through an \emph{in-order delay--throughput} tradeoff. \emph{In-order delay} is the end-to-end time from message generation until information is delivered in the order messages were generated, while \emph{throughput} is the rate at which information is successfully delivered to the receiver, accounting for protocol overhead and losses. The tradeoff arises because pushing for higher throughput and reliability often increases queueing and ordering stalls, raising in-order delay, while reducing in-order delay often requires added redundancy (e.g., using error-correction), which can reduce effective throughput.
In connected multi-robot systems, where both timeliness and reliability are essential, in-order delay is therefore the notion of delay that best reflects what the autonomy pipeline actually experiences.

Along the communication stack, several factors degrade these metrics: path-loss, fading, and obstruction can cause packet erasures; medium access and contention can increase channel access time; buffers in intermediate nodes add queueing delay or overflow under high load; and multi-hop forwarding and recovery mechanisms (e.g., feedback-based retransmissions using automatic repeat request (ARQ)) can further increase delay and reduce effective throughput. In multi-robot systems, these effects primarily manifest as loss, in-order delay, and variability in delivery time.


\subsection{Transport Protocols}
Among the outlined layers, we focus on the transport layer, where end-to-end delivery protocols are commonly implemented and which directly serves the autonomy algorithms. 
To study how delivery behavior affects the autonomy pipeline, we describe three transport protocols that represent common delivery behaviors with distinct timing characteristics: best-effort delivery, retransmission-based reliability with in-order delivery, and adaptive coded transport. These protocols determine the timing and availability characteristics of packet delivery that drive the estimator in cooperative localization and the reliable delivery required by the abort-by-deadline mechanism in the overtaking case study.

\vspace{0.2cm}
\noindent \emph{A.1) Classical Non-Coded Transport-Layer Protocols}:
\vspace{0.2cm}

The \emph{User Datagram Protocol}  (UDP)~\cite{postel1980udp} serves as our unreliable best-effort baseline.
UDP provides a minimal, connectionless transport protocol without acknowledgments, retransmissions,
congestion control, or ordering guarantees. Its simplicity results in low protocol overhead and
minimal added delay, but lost or out-of-order packets are not recovered. UDP is therefore a common
choice when best-effort delivery is sufficient and the application can tolerate occasional loss or
reordering. 

\emph{Selective-Repeat ARQ} (SR-ARQ)~\cite{weldon1982improved} serves as our retransmission-based
reliable baseline. SR-ARQ improves reliability by retransmitting only packets that are explicitly reported as missing. Because it mitigates unnecessary retransmissions, SR-ARQ is highly throughput-efficient and is therefore commonly used as a baseline for reliable data transfer.
SR-ARQ enforces ordered delivery using a sliding window of outstanding packets, which allows multiple packets to be in transmission at once, but the receiver releases them to the application only in order, even if later packets arrive first. Under loss, this ordered release can increase in-order delay due to head-of-line blocking.


\vspace{0.2cm}
\noindent \emph{A.2) Code-Based Transport-Layer Protocols}:
\vspace{0.2cm}

Next, we consider a \emph{network coding}-based transport protocol as an alternative to feedback-driven retransmissions. In our setting, adaptive coded transport is relevant because it can reduce in-order delivery stalls under loss and improve the timeliness of information delivered to the autonomy pipeline.

Network coding improves robustness by transmitting coded packets that combine information from multiple original packets, rather than sending or retransmitting a specific missing packet. A common realization of this idea is \emph{Random Linear Network Coding (RLNC)}~\cite{Ho2006RLNC}, which serves as the foundation for the \emph{adaptive causal} coded protocol (e.g., AC-RLNC) used later in this work. In RLNC, the receiver can recover the originals once it has collected enough linearly independent coded packets, which makes recovery depend on how many coded packets were received rather than which ones. From the perspective of autonomy, the key distinction between retransmission-based transport and coded transport lies in how losses are repaired: ARQ-based schemes recover specific missing packets after feedback, coupling recovery time to the RTT and potentially inducing head-of-line blocking, whereas network coding transforms erasure recovery into a degree-of-freedom accumulation problem, where any sufficiently large set of independent coded packets enables decoding. This decouples reliability from the identity of individual packets and reduces sensitivity to burst losses (i.e., multiple consecutive packet erasures) and timing variability, properties that are particularly relevant for real-time multi-robot cooperation.

\emph{Adaptive Causal RLNC (AC-RLNC)}~\cite{9076631} serves as our adaptive coded transport protocol.
It extends RLNC for real-time settings (i.e., delay-sensitive applications where messages must be delivered within tight deadlines) 
by using a sliding coding window and adapting redundancy
both proactively (to mask expected losses) and reactively (based on feedback). This enables robust
and timely delivery without waiting for specific missing packets: once enough coded packets are
received, the receiver can decode the corresponding set of original packets and release them in
order. 
In particular, by reducing in-order stalls while maintaining reliability, AC-RLNC helps mitigate and manage the in-order delay--throughput tradeoff that is central to our setting.

\subsection{Communication Challenges in Multi-Robot Systems}
The above challenges are exacerbated in MRS.
Robots operate in cluttered, dynamic environments where motion and obstacles cause time-varying degradation in link quality, while the robots seek relevant (freshest) task-oriented data delivered on a tight schedule. Teams share wireless spectrum, leading to contention that induces unpredictable timing variability in packet delivery. Topology changes as robots move in and out of range,
frequently altering which communication links exist. In addition, practical deployments are
often \emph{heterogeneous}: different agents may experience different link qualities, traffic
loads, and communication modalities, so end-to-end behavior can vary across the team and over
time. Extensive reviews highlight that such factors make multi-robot communication considerably
more complex than static wireless networking, and that communication mechanisms are still rarely
co-optimized for robotic task requirements~\cite{Gielis2022Review,10418581}.


Crucially, in these settings we do not necessarily care about delay and throughput in isolation.
What matters is whether delivered information is still useful to the autonomy pipeline. We
therefore distinguish \emph{urgency}, where a message is useful only if it arrives before a
task-dependent deadline, and \emph{freshness}, where an old update may be stale even if it is
eventually delivered. These considerations motivate the case-study analysis presented in this
work and point to the need for communication mechanisms that better align delivery behavior with
autonomy requirements.


\section{System Models and Case Studies}
\label{sec:sysmodel}

Here,  we describe the two system models considered in our case studies: 
(i) a multi-robot cooperative localization problem subject to communication delay and packet loss, and 
(ii) a safety-critical overtaking scenario in which communication timing determines collision avoidance. 
We formalize the motion, sensing, communication, transport-layer, and estimation models used in both studies. 
In both settings, we sought to design scenarios that are on the one hand simple to describe, yet capture important problems using realistic assumptions and models.

\subsection{Cooperative Localization 
}\label{subsec:coopL}
This setting consists of a group of $n$ robots moving within a shared workspace in a semi-random fashion (Fig.~\ref{fig:system_view}; for visual clarity, the figure depicts a simplified example with five robots). The goal of each robot is to estimate its state by relying on information from two noisy sources: (1) self-measurements using a GPS-like sensor and (2) inter-robot measurements generated by neighboring robots and exchanged over the wireless channel, subject to communication-induced delay and loss.

\label{subsec:cooploc_model}
\niceparagraph{Motion model.}
Each robot $i\in\mathcal{R}$ evolves at discrete time slots $t\in\{0,\dots,T\}$ with sampling period $\Delta t$ and state $\bm{x}
^i_t = (
    x^i_t, y^i_t, \theta^i_t
    )
    \in \mathbb{R}^3$,
where $(x^i_t,y^i_t)$ denotes the planar position and $\theta^i_t$ denotes the heading. The robot follows  Ackermann steering kinematics \cite{lavalle2006planning} with control input
$   \bu^i_t =(v^i_t,\delta^i_t)$, 
where $v^i_t$ is the forward velocity and $\delta^i_t$ is the steering angle, and wheelbase $L$. 
The deterministic motion update rule is
\begin{equation}
    \bm{x}^i_{t} = f(\bx^i_{t-1},\bu^i_t)
    =
    \begin{bmatrix}
        x^i_{t-1} + \Delta t\, v^i_t \cos\theta^i_{t-1}\\
        y^i_{t-1} + \Delta t\, v^i_t \sin\theta^i_{t-1}\\
        \theta^i_{t-1} + \Delta t \frac{v^i_t \tan(\delta^i_t)}{L}
    \end{bmatrix}
    \label{eq:ackermann}
\end{equation}

The robots are initialized with random positions and accelerations drawn from uniform distributions within predefined bounds, and with zero initial steering angle. During the simulation, control inputs are updated in a piecewise-constant manner: every $\tilde{T}$ \off{\anil{$\tilde{T}$, it was $T$ which was already defined as the time horizon, so I changed it} \ale{ok}} time slots we resample the acceleration and sample a new steering angle, and every $\hat{T}$ time slots we reset the steering angle to zero to avoid persistent circular motion. To prevent collisions, we apply a simple reactive rule independent of the communication mechanism: when two robots are detected to be closing on each other, both robots resample steering commands to increase separation. The velocity is updated by integrating acceleration with saturation to a predefined maximum speed before applying the model in Eq.~\eqref{eq:ackermann}.


\niceparagraph{Sensor model.}
Each robot $i$ obtains two measurement types. The first is a GPS-like \emph{self-measurement} of its location with Gaussian noise $\beps^i_{\mathrm{GPS}}(t)\sim\mathcal{N}(0,\sigma_{\mathrm{GPS}}^2 I_2)$, i.e.,  
$
    \bz^{i,\mathrm{GPS}}_t 
    =
    \begin{bmatrix}
        x^{i, \mathrm{true}}_t \\ y^{i, \mathrm{true}}_t
    \end{bmatrix}
    + \beps^i_{\mathrm{GPS}}(t)$.

This measurement can be obtained by the robot at each time slot $t$, where $\sigma_{\mathrm{GPS}}$ is a fixed standard deviation parameter of the GPS measurement noise, assumed known to the robot. 
In addition, robot $i$ can obtain an \emph{inter-robot measurement} (e.g., by using a light detection and ranging (LiDAR) sensor)  of robot $j$ as 
\begin{equation*}
    \bz^{ij}_t 
    = 
    \begin{bmatrix}
        x^{j, \mathrm{true}}_t \\ y^{j, \mathrm{true}}_t
    \end{bmatrix}
    + \beps^{ij}_{\mathrm{L}}(t), 
    \quad \text{for} \quad
    \beps^{ij}_{\mathrm{L}}(t)
    \sim 
    \mathcal{N}(0,\sigma_{\mathrm{L}}^2 I_2),
\end{equation*}
i.e., $\bz^{ij}_t$ is robot $i$'s noisy estimate of robot $j$'s true global position, where the additive term $\beps^{ij}_{\mathrm{L}}(t)$ captures both sensing noise and the additional uncertainty induced by the measuring robot's pose uncertainty when expressing the detection in global coordinates. Accordingly, we parameterize the measurement-noise standard deviation
    $\sigma_{\mathrm{L}}
    = \sigma_{\mathrm{internal}}
    + {d_{ij}(t)}/{\sqrt{2}\,M}$,
where $d_{ij}(t)$ is the Euclidean distance between robots $i$ and $j$, and $\sigma_{\mathrm{internal}}$ is a fixed constant that captures additional uncertainty introduced by representing the inter-robot observation as a noisy global-position measurement (rather than using an explicit range-and-bearing model). The term $\sqrt{2}\,M$ equals the workspace diagonal length (with $M$ the side-length of the square workspace) and serves as an upper-bound range scale, so $\sigma_{\mathrm{L}}$ increases with distance. The value of $\sigma_{\mathrm{L}}$ is not assumed to be known to the robots.

\niceparagraph{Communication channel model.} 
The goal of the communication layer in this setting is not high throughput per se, but timely and reliable delivery of relative measurements that directly affect state estimation accuracy.
Communication between robots is modeled as point-to-point communication between robot pairs. Specifically, when a robot sends an inter-robot measurement to another robot, the packet traverses the corresponding wireless link, modeled as a Binary Erasure Channel (BEC), where packets are independently erased with probability $\epsilon\geq 0$ and otherwise delivered after a fixed one-way delay.
This abstraction captures packet drops due to fading, interference, or contention, while keeping the model analytically simple.
Feedback is provided via a backward acknowledgment channel, i.e., by sending acknowledgment, ACK, or a negative-acknowledgment, NACK, messages according to the erasure realizations. 
For simplicity, we assume the backward channel is reliable and experiences the same one-way delay~\cite{9076631}.


Recall that $\mathrm{RTT}$ is the round-trip time between two robots (assumed constant in our model). 
A successfully delivered packet becomes available for estimation after a one-way delay of $\mathrm{RTT}/2$ (a standard modeling simplification). 
When un-coded retransmission-based protocols are used, loss recovery requires waiting for feedback over a full $\mathrm{RTT}$, which directly impacts in-order delivery delay. 

Each packet consists of the sender ID, target ID, 
a synchronized timestamp of the measurement, and the measurement content, which includes robot $i$'s LiDAR-based estimate of robot $j$'s global position together with a noisy range measurement $\hat d_{ij}(t)=d_{ij}(t)+\epsilon_{d,ij}(t)$, where $\epsilon_{d,ij}(t)\sim\mathcal{N}\!\left(0,\left({d_{ij}(t)}/{\sqrt{2}M}\right)^2\right)$ (used for covariance construction).
In general, in the considered communication model, packets may arrive late, out-of-order, or be erased permanently.


\niceparagraph{Transport protocols.}
We consider the three transport mechanisms introduced in Sec.~\ref{sec:background}:
(1) \emph{UDP}, without retransmission or ordering guarantees. (2)  \emph{Selective Repeat ARQ}, with reliable transmission via selective packet retransmission and a sliding window of size $ W_{\mathrm{SR}} = a \beta\, \mathrm{RTT}$,  which can stall if the base packet is repeatedly lost. (3) \emph{AC-RLNC}, which uses a sliding coding window and adaptive redundancy (via a-priori and post-priori forward error correction (FEC)). The maximum coding window size is 
$
    W_{\mathrm{AC}} = b\,\beta\,\mathrm{RTT}
$. 
Here, $a$ and $b$ are tunable parameters selected to manage the in-order delay--throughput tradeoff. We choose $a$ so that the SR-ARQ window remains sufficiently large to keep the sender pipeline full under RTT-scale feedback, and $b$ to control the coding span in AC-RLNC and thereby limit the potential decoding delay.
The communication rate is
\begin{equation*}
    f_c = \beta f_s,
    \quad \text{for} \quad
    \beta =
    \bigg\lceil
    \frac{1}{\max\{1 - \epsilon - \alpha,\,\lambda\}}
    \bigg\rceil,
\end{equation*}
where $f_s = 1/\Delta t$ is the simulation update rate, $\alpha$ is a small safety margin when computing $\beta$, and $\lambda>0$ is a lower bound that prevents the computed rate from becoming excessively large under very high erasure probabilities. 
This adaptive rate ensures that transmission remains compatible with the effective channel capacity, maintaining stability even under high packet-loss conditions.  
When network quality deteriorates (i.e., $\epsilon$ increases), $\beta$ increases accordingly, enabling more frequent packet transmissions or coded-packet repetitions.  
AC-RLNC benefits in particular from this mechanism, as it uses the additional transmission opportunities to send coded packets. Each such transmission contributes a degree of freedom at the receiver, increasing the likelihood of earlier decoding and thereby reducing the in-order delivery delay.

\niceparagraph{State estimation.}
Each robot maintains an extended Kalman filter (EKF) to estimate its state from noisy self- and inter-robot measurements. The applied control $\bar{u}^i_t$ during the prediction step is modeled as a noisy measurement of the true input,
    $\bm{\bar{u}}^i_t:=\bu^i_t + \bm{\eta}^i_t$,  for $\bm{\eta}^i_t \sim \mathcal{N}(0,\Sigma_u)$,
where $\Sigma_u=\mathrm{diag}(\sigma_v^2,\sigma_\delta^2)$ is a fixed covariance whose value is not known to the robots.
The prediction of EKF is obtained from 
    $\bm{\hat{x}}^i_{t} = f(\hat{x}^i_{t-1},\, \bar{u}^i_t)$,
    and 
    $P^i_{t} = F^i_t P^i_{t-1} (F^i_t)^\top + Q$,
where $F^i_t$ is the Jacobian of the motion model, and the process-noise covariance is 
$
    Q = \mathrm{diag}\!\left(
        \sigma_{\mathrm{process}}^{2},\;
        \sigma_{\mathrm{process}}^{2},\;
        \sigma_{\theta}^{2}
    \right),
$
with $\sigma_{\mathrm{process}}$ and $\sigma_{\theta}$ denoting fixed standard deviation parameters of the translational and heading process uncertainties, assumed known to the robot.
In addition, 
each received measurement $\bz$ (either a GPS self-measurement $\bz^{i,\mathrm{GPS}}_t$ or an inter-robot measurement $\bz^{ij}_t$ delivered over the communication channel) triggers the standard EKF update
\begin{align*}
    & K = P^- H^\top (H P^- H^\top + R)^{-1},\\
    & \bm{\hat{x}}^+ = \bm{\hat{x}}^- + K(\bz - H\bm{\hat{x}}^-), \text{ and }
    P^+ = (I-KH)P^-,
\end{align*}
where $H$ denotes the observation Jacobian and $R$ the measurement covariance. Note that, the
robots do not know the true distribution from which the LiDAR measurements are drawn in the considered setting. I.e.,  the  inter-robot measurements are generated using the
baseline LiDAR uncertainty $\sigma_{\mathrm{L}}$, but each robot constructs its own
measurement covariance using its predicted process uncertainty. For GPS updates we use
$
R_{\mathrm{GPS}} = \sigma_{\mathrm{GPS}}^{2} I_2,
$
while for inter-robot measurements the robot approximates the covariance as $R_{\mathrm{LiDAR}}(t)
= (\sigma_{\mathrm{process}} + {\hat d_{ij}(t)}/{\sqrt{2}M})^2 I_2$, where $\hat d_{ij}(t)$ is the (noisy)  range estimate obtained from the LiDAR detection and transmitted to the receiving robot. This reflects the estimator's assumption that measurement uncertainty increases with range, with an additional offset term ($\sigma_{\mathrm{process}}$) that captures the effect of the measuring robot's pose uncertainty on the transmitted global-position measurement. 


\niceparagraph{Delay-handling mechanisms.}
Packet delays cause measurements to arrive with information age $d := t - \tau$, where $t$ is the current time slot and $\tau$ is the time slot at which the measurement was generated. Two approaches are considered.
The \emph{naive time-window} approach processes each measurement immediately if its age satisfies $d \le D$, where $D$ is the estimator's delay-handling window size, and discards it otherwise, causing delayed inter-robot measurements to be applied out of sequence and potentially degrading estimation accuracy. As a more robust alternative, we introduce the \emph{iterative re-estimation}  (I-ReE) approach. 
I-ReE is a communication-aware method that maintains a sliding window of the past $D$ states, covariances, controls, and received measurements. When a delayed measurement with age $d \le D$ arrives, it is inserted into the appropriate buffer, and the EKF is replayed from $t-d_{\max}$ to $t$, where $d_{\max}$ is the largest active delay. By reprocessing the prediction and update steps in temporal order, I-ReE incorporates delayed information consistently and avoids the degradation caused by out-of-sequence updates. 




\subsection{Safety-Critical Overtaking Under Timing Constraints}
\label{subsec:overtake_model}

We consider a safety-critical overtaking scenario in which an ego vehicle must decide whether to abort an ongoing overtake based solely on the timeliness of V2V packet reception (Fig.~\ref{fig:overtake_overview}). All vehicle motion here is deterministic; uncertainty arises exclusively from packet delays and communication channel erasures.


The scenario involves the ego vehicle $A$, a slower truck $T$, and an
oncoming vehicle $B$. Both $A$ and $B$ travel on the outer lane but in opposite
directions, while $T$ travels on the inner lane in the same direction as $A$. All
vehicles follow the Ackermann steering model (Eq.~\eqref{eq:ackermann}).
The ego vehicle begins the maneuver already travelling in the outer lane behind $T$ (Fig.~\ref{fig:overtake_overview}).
During the overtake, $A$ remains in this lane while $B$ approaches head-on in the same
lane. If $A$ does not abort the maneuver early enough, the two vehicles may collide. 
Note that no direct sensing of $B$ is
assumed by vehicle $A$ (except in late situations where a collision is imminent), as vehicle $T$ obstructs its view due to the problem geometry. Instead, the ego vehicle must rely entirely on receiving a sufficient number of packets
to infer that $B$ is approaching in its lane and that the overtake must be aborted. 


To determine whether the ego vehicle can still safely abort the overtake given the ego vehicle's decisions, the considered setting
rolls out an abort maneuver beginning at a candidate time slot: a strong braking with deceleration $a_{\max}$ followed by a smooth,
bounded steering input $\delta_{\mathrm{abort}}$ that guides the vehicle from the outer
lane back toward the inner lane behind the truck. The steering profile produces a fast
and stable inward trajectory without excessive lateral overshoot. During the rollout,
full oriented-bounding-box collision checks against both the truck and the oncoming
vehicle are performed. A candidate time slot is marked safe if ego vehicle $A$  completes the abort
without collision; the latest such time slot defines the abort deadline.

\off{
\begin{figure*}[t]
    \centering
    \includegraphics[trim={4cm 0 3cm 0},clip,width=\textwidth]{images/simulations/ArcOvertakeSim_sim_overview.pdf}
    \caption{Initial configuration of the overtaking scenario. Ego vehicle $A$ follows the outer lane behind truck $T$, while oncoming vehicle $B$ approaches from the opposite direction on the same lane. Timely V2V packet reception is required for $A$ to detect the oncoming hazard and abort. The red status band reports the current slot index, the number of received packets, the computed abort deadline (by which the required packets must be received to allow a successful abort), the abort trigger (indicating when sufficient packets have been received and the abort maneuver is initiated), and the instantaneous speeds of ego vehicle $A$ and truck $T$. \kiril{Even when stretched on two columns, the figure is still hard to read. It could be worth zooming in to keep the portion of the road on which all the robots are visible.} \anil{Done}}
    \label{fig:overtake_overview}
\end{figure*}
}

\niceparagraph{Communication and information model.}
To determine whether to abort the overtake maneuver, the ego vehicle relies on messages received from vehicle $B$ via an exchange of generic V2V packets over the same delayed binary-erasure
channel introduced in Sec.~\ref{subsec:cooploc_model}. These packets represent the minimal information needed to
establish a communication session and convey situational context (e.g., presence of an
oncoming vehicle, maneuver intent, or status metadata).  

Vehicle $A$ considers the number $N(t)$ of packets successfully received by time $t$ from $B$, and aborts if and only if $N(t)$ exceeds a certain threshold $\text{msg}_{\mathrm{req}}$ before the abort deadline arrives.  
%
To emulate realistic V2V behavior along the road, packet loss varies over time in a
piecewise-constant manner. The time slot horizon is divided into $J$ contiguous intervals (such that ${T}/{T'}=J \in \mathbb{N}$), each
assigned a fixed average erasure probability $\epsilon(t')$ for $t'\in\{1,\ldots,J\}$. These probabilities decrease across intervals,
reflecting the improving channel conditions as $A$ and $B$ move closer. Packet successes
in each interval follow a Bernoulli process with success probability $1-\epsilon(t')$. This design captures the abrupt changes in
connectivity commonly observed when vehicles move between areas of weak and strong
coverage. Finally, we mention that this scenario uses the same reliable transport mechanisms described in
Sec.~\ref{sec:background} and Sec.~\ref{subsec:cooploc_model}, namely SR-ARQ and
AC-RLNC, while UDP is omitted since it provides no recovery mechanism for the required packet delivery before the abort deadline. The protocol behavior here impacts only the arrival pattern of packets (i.e., their in-order delivery delay and throughput). 

\begin{figure*}[t]
    \centering
    \subfloat[Naive vs.~communication-aware approaches. Note that the blue curve overlaps with the orange curve.]{
        \includegraphics[width=0.465\textwidth]{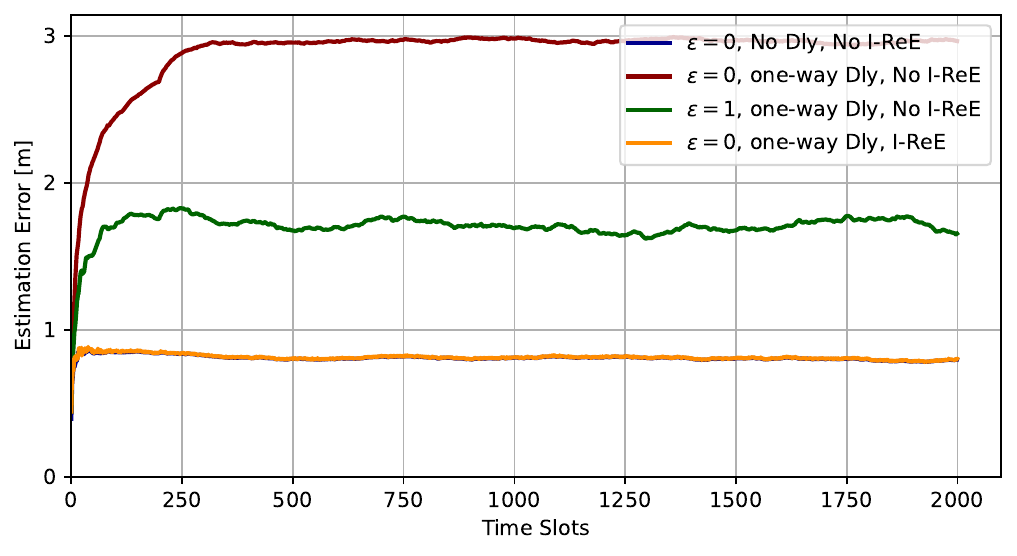}
        \label{fig:sub1}
    }
    \hfill
    \subfloat[Protocol comparison under erasures and delay.]{
        \includegraphics[width=0.48\textwidth]{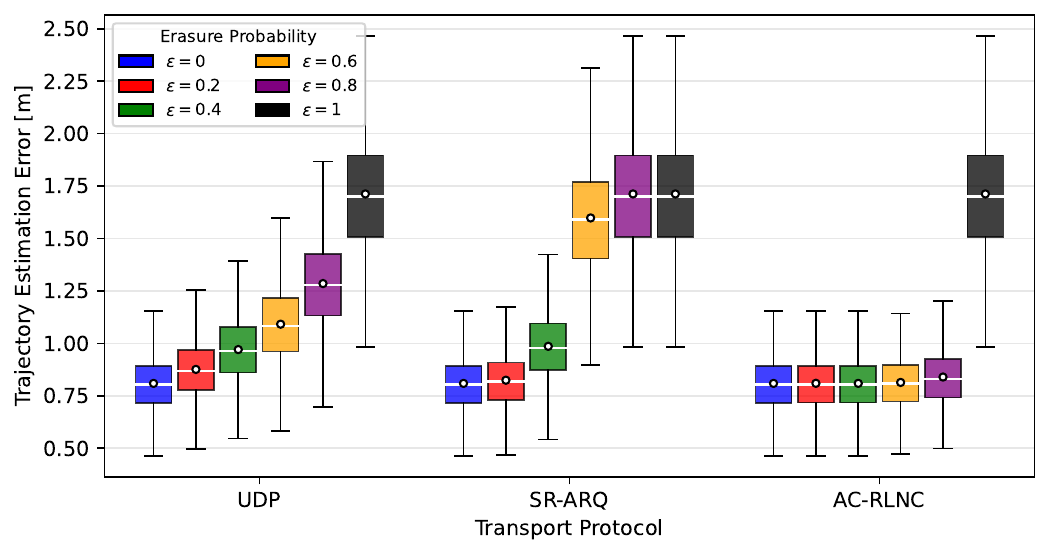}
        \label{fig:sub2}
    }
    \caption{Cooperative localization estimation error under different communication conditions, transport protocols, and delivery delays. Here $\epsilon$ denotes the packet erasure probability. ``No Dly'' denotes $\mathrm{RTT}=0$, while ``one-way Dly'' denotes a fixed one-way delay of $\mathrm{RTT}/2$. ``No I-ReE'' corresponds to the naive approach that updates the EKF upon packet arrival, while ``I-ReE'' denotes the delay-aware iterative re-estimation method. In (a), curves without a protocol name correspond to the estimator-only comparison under the stated delay/erasure condition. In (b), each box summarizes the per-time-slot terms of $\mathrm{Err}(T)$ over the final post-transient window for one protocol and $\epsilon$; the horizontal line indicates the median and the circle indicates the mean. For reliable protocols, packets additionally experience protocol-dependent in-order delivery delay due to retransmissions (SR-ARQ) or decoding (AC-RLNC).}
    \label{fig:cl_fig}
    \vspace{-17pt}
\end{figure*}

\section{Experimental Setup and Results}\label{sec:experiments}
We evaluate how communication delay, packet loss, and protocol behavior affect (i) distributed estimation accuracy in a multi-robot localization scenario and (ii) decision-making reliability in a safety-critical high-speed overtaking scenario. Full implementation can be found in \href{https://github.com/AnilZaher/network-coding-mrs}{github.com/AnilZaher/network-coding-mrs}. 
\iftoggle{arxiv}
{
The parameters used in the experiments were set to reflect realistic behavior and models, and are 
reported in Table~\ref{tab:cl_params} and~\ref{tab:overtake_params}. 
}
{
The parameters used in the experiments were chosen to reflect realistic behavior and model characteristics\footnote{Parameter values. \textbf{Localization:} $n=10$, $M=200$ m, $\Delta t=0.1$ s, $T=2000$ slots, $\tilde{T}=8$ slots, $\hat{T}=3$ slots, $L=2.5 $ m, $\sigma_{\mathrm{GPS}}=3$ m, $\sigma_{\mathrm{internal}}=2$ m, $\sigma_{\mathrm{process}}= 1$ m, $\sigma_{\theta}=1^\circ$, $\sigma_{v}=3$ m/s, $\sigma_{\delta}=2^\circ$, RTT$=$ 4 slots, $D=10$ slots, $f_c=\beta f_s,\; f_s = 1/\Delta t$, $\alpha=0.11$, $\lambda=0.15$. \textbf{Overtake:} Lane width  $= 3.5$ m, ego and oncoming vehicle velocity $=28$ m/s, truck velocity $22$ m/s, $a_{\max}=10$ m/s$^2$, $\delta_{\mathrm{abort}}=10^\circ$, $\Delta t=0.05$ s, $T=160$ slots, $T'=16$ slots, $\mathrm{RTT}=8$ slots, and $1-\epsilon(t') \in [0.1, 0.9]$. In both settings, $W_{\mathrm{SR}}=a\beta\,\mathrm{RTT}$ and $W_{\mathrm{AC}}=b\beta\,\mathrm{RTT}$ with $a=2$ and $b=1.5$; for overtake, $\beta=1$.}.
}



\subsection{Cooperative Localization Results} \label{subsec:cooploc_results}
We quantify estimation performance using a trajectory-estimation metric computed over $k(t)$ time slots:
\[
\mathrm{Err}(t)
=
\frac{1}{n}
\sum_{i=1}^{n}
\frac{1}{k(t)}
\sum_{\tau=t-k(t)+1}^{t}
\lVert \bm{\hat{x}}^i_\tau - \bx^{i, \mathrm{true}}_\tau \rVert_2 .
\]
This metric averages the recent position-estimation deviations across all robots, reducing high-frequency fluctuations while preserving the temporal evolution of the estimator.
For the estimator-only comparison in Fig.~\ref{fig:sub1}, we use $k(t)=\min(200,t+1)$. For the protocol comparison in Fig.~\ref{fig:sub2}, we evaluate the same metric at the final time $T$ with $k(T)=T-100+1$, so the box plot summarizes how the final estimate of the traversed trajectory compares with the ground truth under each protocol and erasure probability. This final-time behavior serves as a proxy for the entire run, since the estimates are relatively steady over time. 
We report results for a single seeded realization of the randomized robot motions (and measurement noise), using the same seed across all protocol comparisons; varying the seed produced similar trends and does not materially change the conclusions.

We evaluate three communication regimes: 
(i)~\emph{Ideal communication}, i.e., no delay ($\text{RTT}=0$) and no packet loss ($\epsilon=0$);  
(ii)~\emph{Delayed communication}, i.e., lossless delivery with only one-way delay 
and no packet loss;  
(iii)~\emph{Lossy with delay}, inter-robot measurements are subject to both  delay and erasure 
probability $\epsilon$.  
When there is no delay (ideal case or full erasure), the I-ReE estimator reduces to the 
standard EKF as no out-of-sequence measurements occur. Fig.~\ref{fig:cl_fig} presents the resulting estimation error trajectories.

In the first set of experiments (Fig.~\ref{fig:sub1}), we evaluate the effect of ideal and delayed communication, coupled with a naive or a communication-aware estimation approach (I-ReE). The ideal case ($\epsilon = 0$, No Dly, No I-ReE) serves as the baseline
and reflects the best achievable estimator performance. In this regime, I-ReE collapses
to the naive EKF because no measurements arrive delayed or out of sequence. 
When a one-way delay is introduced without packet loss
($\epsilon = 0$, one-way Dly, No I-ReE), the naive EKF exhibits a marked increase in estimation
error due to out-of-order incorporation of delayed measurements. The corresponding I-ReE
curve ($\epsilon = 0$, one-way Dly, I-ReE) remains close to the baseline by restoring
chronological consistency before applying updates. In the lossy condition ($\epsilon = 1$, one-way Dly, No I-ReE), all inter-robot measurements are
erased, so no delayed information ever arrives, and I-ReE again reduces to the naive EKF. Overall, this comparison shows that delay, rather than loss, is the key factor
that motivates the need for a delay-aware estimator: whenever delayed measurements are
present, I-ReE provides a substantial improvement. 
Next, we assess the impact of the transport protocols under erasures and delay, combined with the communication-aware estimator (Fig.~\ref{fig:sub2}).
In all cases, inter-robot measurements are subject to both packet erasures and protocol-dependent delivery delay.
UDP exhibits monotonic degradation as erasures increase, since lost measurements are 
never recovered.  
SR-ARQ performs well under moderate erasure but degrades sharply at high loss due to 
retransmission stalls that delay packets beyond the usable estimation window.  
AC-RLNC maintains near-ideal performance across all tested erasure probabilities below~1,
owing to its adaptive coded redundancy and steady delivery of decodable packets.


Overall, the findings here show that delay is the dominant factor impairing cooperative localizationand that traditional protocols face structural limitations due to either loss (UDP) or latency stalls (SR-ARQ).  The coding solution AC-RLNC combined with I-ReE robustly preserves near-ideal performance in the scenarios tested. This highlights the value of aligning transport behavior with estimator design and motivating tighter communication and estimation co-design.

\subsection{Safety-Critical Overtaking Results}
\label{subsec:overtake_results}
The overtaking simulation evaluates whether timely V2V packet delivery allows the ego
vehicle to abort the maneuver before the physically computed abort deadline. The deadline is obtained from the abort rollout described in Sec.~\ref{subsec:overtake_model}. For this scenario and problem parameters, the abort deadline is at time slot $t=110$.
We require
$
\text{msg}_{\mathrm{req}} = 25
$
packets to be successfully received,
representing a conceptual 
requirement to establish a communication session and convey situational information (e.g., heading)  
used by the abort logic, consistent with V2X collision-prediction systems that use sequences of received vehicle-state updates for safety decisions~\cite{s23031260}.



\begin{figure}
\centering
\iftoggle{arxiv}{
\vspace{5pt} \includegraphics[width=0.95\columnwidth]{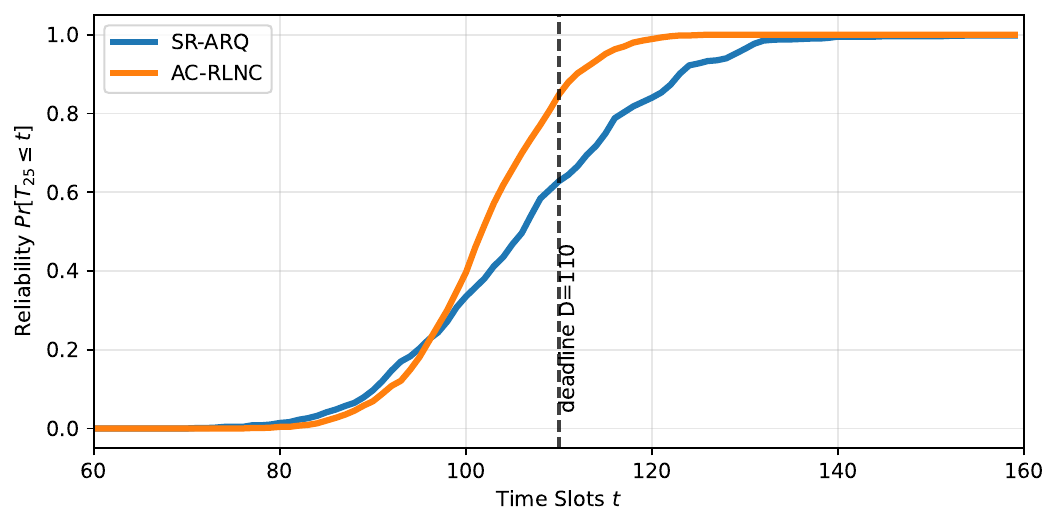} \vspace{-10pt}
}
{
\vspace{5pt} \includegraphics[width=0.82\columnwidth]{images/plots/reliability_latency.pdf} \vspace{-10pt}
}
    \caption{Overtaking reliability--latency function $\Pr[T_{25} \le t]$, where $t$ (horizontal axis) denotes time slots and $T_{25}$ is the arrival time of the 25th successfully received packet. The value at $t=110$ corresponds to the probability of satisfying the abort-by-deadline requirement.\off{\anil{Will update the x-axis label to time steps t,in a bit}}}
    \label{fig:reliability_latency}
    \vspace{-20pt}
\end{figure}

For each protocol, we examine the reliability-latency curve $
\Pr[T_{25} \le t]$, 
where $T_{25}$ denotes the arrival time (in time slots) of the 25th successfully received packet.  
This curve captures the full distribution of completion times under the 
time-varying erasure process. 
We estimate it by a Monte-Carlo test from $N=1000$ independent simulation runs; across runs, packet delivery differs due to different realizations of the time-varying packet erasures.
Fig.~\ref{fig:reliability_latency} shows the results.
In the scenario tested, AC-RLNC achieves a rapid rise in reliability, indicating that the required packets typically arrive much earlier than the deadline.  
SR-ARQ, by contrast, exhibits a long latency tail due to retransmission stalls, which significantly increase the probability of late completion. Evaluating the curve at $t=110$ yields the probability of meeting the abort deadline for each protocol, showing a higher completion probability for AC-RLNC (about $80\%$) than for SR-ARQ (about $60\%$). A typical outcome is illustrated in Fig.~\ref{fig:overtake_overview}.  This result attests again to the strength of the coded solution.




\section{Discussion and Future Work}
\label{sec:conclusion}
Our case study shows that communication delay, loss, and transport behavior can directly affect both estimation accuracy and safety-critical decision making in connected multi-robot systems. In cooperative localization, delayed and out-of-sequence measurements degrade accuracy when fused in arrival order, while delay-aware re-estimation mitigates this effect; combined with adaptive coded transport, accurate trajectory estimates are maintained even under impaired links. In the overtaking scenario, safety reduces to meeting a strict abort deadline: retransmission-driven latency tails increase deadline misses, whereas adaptive coded delivery significantly improves the probability of triggering an abort in time in the scenarios tested. 
Overall, the results reinforce a central point: autonomy pipelines and communication mechanisms should be co-designed, since protocol-induced timing behavior can determine whether information remains usable and whether safety constraints are met. 

Building on these insights, future work should formalize task-level information requirements such as deadlines, freshness, relevance, and prioritization, and use them to guide both estimator/controller design under delayed updates and communication mechanisms and new streaming coding that schedules and allocates redundancy according to data timeliness and task criticality. The longer-term goal is principled co-design methods that enable reliable MRS behavior under heterogeneous and time-varying networks.

\niceparagraph{Acknowledgments.} 
The AI system ChatGPT was used for light editing and grammar enhancement, a preliminary literature review, and aesthetic enhancements in Fig.~\ref{fig:system_view}. 
\bibliographystyle{IEEEtran}

\bibliography{bibtex/bib/refs.bib}


\iftoggle{arxiv}
{
\newpage
\appendix
%
%
\begin{table}[!ht]
    \centering
    \caption{Cooperative Localization Simulation Parameters}
    \label{tab:cl_params}
    \begin{tabular}{l c c}
        \hline
        \textbf{Parameter} & \textbf{Symbol} & \textbf{Value} \\
        \hline
        Number of robots & $n$ & 10 \\
        Workspace side length & $M$ & 200 m \\
        Slot duration & $\Delta t$ & 0.1 s \\
        Horizon & $T$ & 2000 time slots \\
        Control update period & $\tilde{T}$ & 8 time slots \\
        Steering reset period & $\hat{T}$ & 3 time slots \\
        Wheelbase & $L$ & 2.5 m \\
        GPS noise std. & $\sigma_{\mathrm{GPS}}$ & 3 m \\
        Baseline LiDAR uncertainty & $\sigma_{\mathrm{internal}}$ & 2 m \\
        Process noise (x,y) & $\sigma_{\mathrm{process}}$ & 1 m \\
        Heading noise & $\sigma_{\theta}$ & $1^\circ$ \\
        Control noise (velocity) & $\sigma_{v}$ & 3 m/s \\
        Control noise (steering) & $\sigma_{\delta}$ & $2^\circ$ \\
        Round-trip time & RTT & 4 time slots\\
        Estimation window size & $D$ & 10 time slots \\
        SR-ARQ window & $W_{\mathrm{SR}}$ & $a\beta\,\mathrm{RTT},\; a=2$ \\
        AC-RLNC window & $W_{\mathrm{AC}}$ & $b\beta\,\mathrm{RTT},\; b=1.5$ \\
        Protocol rate & $f_c$ & $\beta f_s,\; f_s = 1/\Delta t$ \\
        Safety margin & $\alpha$ & 0.11 \\
        Denominator floor & $\lambda$ & 0.15 \\
        \hline
        \multicolumn{3}{p{0.95\columnwidth}}{\footnotesize
        \textit{Note:} $\beta$ is the transmission-rate scaling factor defined in
        Sec.~\ref{subsec:cooploc_model} and $\alpha$ is a safety-margin parameter used when computing $\beta$.
        The floor parameter $\lambda$ prevents denominator collapse (and thus rate blow-up) in the $\beta$ computation.
        $\sigma_{\mathrm{internal}}$ reflects a conservative baseline LiDAR uncertainty consistent with worst-case no-communication performance.
        RTT denotes the round-trip time of the communication channel.}
    \end{tabular}
\end{table}

%
\begin{table}[!ht]
    \centering
    \caption{Safety-Critical Overtaking Simulation Parameters}
    \label{tab:overtake_params}
    \begin{tabular}{l c c}
        \hline
        \textbf{Parameter} & \textbf{Symbol} & \textbf{Value} \\
        \hline
        Lane width & $w_{\mathrm{lane}}$ & 3.5 m \\
        Car length & $L_{\mathrm{car}}$ & 4.5 m \\
        Car width & $W_{\mathrm{car}}$ & 1.9 m \\
        Truck length & $L_{\mathrm{truck}}$ & 12.0 m \\
        Truck width & $W_{\mathrm{truck}}$ & 2.6 m \\
        Wheelbase & $L_{\mathrm{wb}}$ & $0.6\,L_{\mathrm{vehicle}}$ \\
        Ego velocity & $v_A$ & 28 m/s \\
        Truck velocity & $v_T$ & 22 m/s \\
        Oncoming velocity & $v_B$ & 28 m/s \\
        Maximum braking & $a_{\max}$ & 10 m/s$^2$ \\
        Abort steering angle & $\delta_{\mathrm{abort}}$ & $10^\circ$ \\
        Slot duration & $\Delta t$ & 0.05 s \\
        Horizon & $T$ & 160 time slots \\
        Round-trip time & RTT & 8 time slots\\
        Required packets & $\text{msg}_{\mathrm{req}}$ & 25 \\
        Deadline & $\text{deadline}$ & 110 time slot\\
        Erasure profile & $1-\epsilon(t')$ & $[0.1,\,0.9]$ \\
        Channel intervals & $T'$ & 16 time slots \\
        Protocol windows & $W_{\mathrm{SR}}, W_{\mathrm{AC}}$ & same as Tab.~\ref{tab:cl_params} \\
        \hline
        \multicolumn{3}{p{0.95\columnwidth}}{\footnotesize
        \textit{Note:} The abort deadline and required packet count correspond to the
        feasibility conditions of the abort maneuver defined in Sec.~\ref{subsec:overtake_model}. 
        The piecewise erasure profile models changing V2V link quality as the vehicles move
        along the trajectory. The protocol window definitions follow those used in the
        cooperative localization experiment (Tab.~\ref{tab:cl_params}), with $\beta = 1$.}
    \end{tabular}
\end{table}

}
{

}



\end{document}